\documentclass{article}
\usepackage{times, graphicx, subfigure, natbib, algorithm, algorithmic, enumitem, float, hyperref, amsmath}

\usepackage[accepted]{icml2017}

\def\f{\frac}\def\l{\left}\def\r{\right}

\icmltitlerunning{Learning from Fewer Samples by In-Domain Transfer}
\begin{document} 
\twocolumn[
\icmltitle{A Brief Study of In-Domain Transfer and Learning from Fewer Samples using A Few Simple Priors}
\begin{icmlauthorlist}
\icmlauthor{Marc Pickett}{goor}
\icmlauthor{Ayush Sekhari}{goores,goo}
\icmlauthor{James Davidson}{goo}
\end{icmlauthorlist}
\icmlaffiliation{goor}{Google Research}
\icmlaffiliation{goores}{Work done as part of the Google Brain Residency program ({\tt g.co/brainresidency}).}
\icmlaffiliation{goo}{Google Brain, Mountain View, CA, USA}
\icmlcorrespondingauthor{Marc Pickett}{pickett@google.com}
\icmlkeywords{transfer learning, machine learning, sample complexity}
\vskip 0.3in
]
\printAffiliationsAndNotice{}

\begin{abstract} 
Domain knowledge can often be encoded in the structure of a network, such as convolutional
layers for vision, which has been shown to increase generalization and decrease
sample complexity, or the number of samples required for successful learning.
In this study, we ask whether sample complexity can be reduced for
systems where the structure of the domain is unknown beforehand, and the structure and
parameters must both be learned from the data.  We show that sample complexity reduction through
learning structure is possible for at least two simple cases.  In studying these cases, we
also gain insight into how this might be done for more complex domains.
\end{abstract} 


\section{Introduction} 
Many domains are constrained by data availability.  This includes domains, such as YouTube
recommendations, which ostensibly have a large amount of data, but which have a long tail of
instances that each have only a handful of data points.  This also includes domains for which humans
require several orders of magnitude less training data than state-of-the-art approaches, such
as motor manipulation, playing Atari, and understanding low-frequency words.
The {\em No Free Lunch Theorem} \cite{wolpert1997no} suggests that a domain prior is needed to
offset sample complexity for these cases.

For domains such as images and audio, {\em convolutions} are commonly used as one such prior.
In our view, convolutions allow for {\em in-domain} transfer by sharing weights among otherwise
weakly-connected areas of the domain, effectively multiplying the number of samples.
For example, if a convolutional layer takes a 28 by 28 MNIST image and uses filters of size 5x5 with
stride 2, this gives 144 5x5 windows per image, which means that for
every image, we have 144 training points for the filter, which effectively decreases our
sample complexity by 144 times for this filter.  The convolutions allow us to internally
transfer information from the top-left 5x5 window to the bottom-right 5x5 window (and to the other 142 windows).

However, there are domains where the structure might not be known beforehand, such as robot
joint angle trajectories or car traffic speed sensor data.  The question we address in
this paper is {\em for domains where we aren't given the structure a priori, is there a weaker
  prior that will both allow a system to learn a structure and leverage the learned structure
  to still have a net decrease in sample complexity over a baseline without the prior?}

We present a simple probability density estimation problem and examine how three
priors affect sample complexity.  We show, at least for one simple domain, the structure can be
recovered merely given the prior that there {\em is} a repeated structure, with the number of
filters and size of windows.  Using this prior, we show how a system may automatically transfer
from parts of the domain where samples are relatively plentiful to parts where samples are more
rare.

\section{The Four Urns Problem} 

\begin{figure*}[ht] 
  \centering{
    \subfigure{(a)
      \includegraphics[width=65mm]{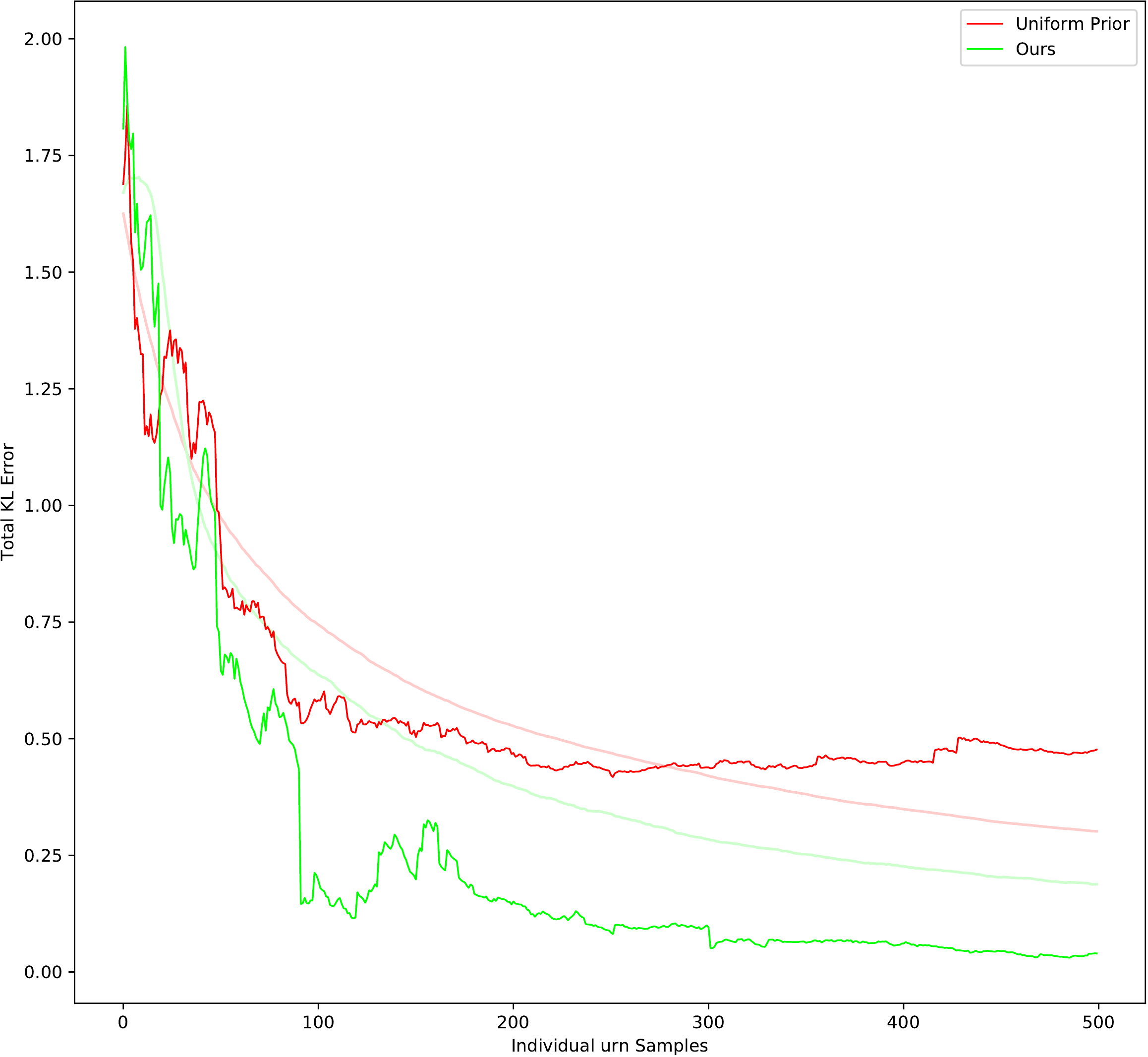}
      \label{fig:foururns1}}
    \hspace{10mm}
    \subfigure{(b)
      \includegraphics[width=65mm]{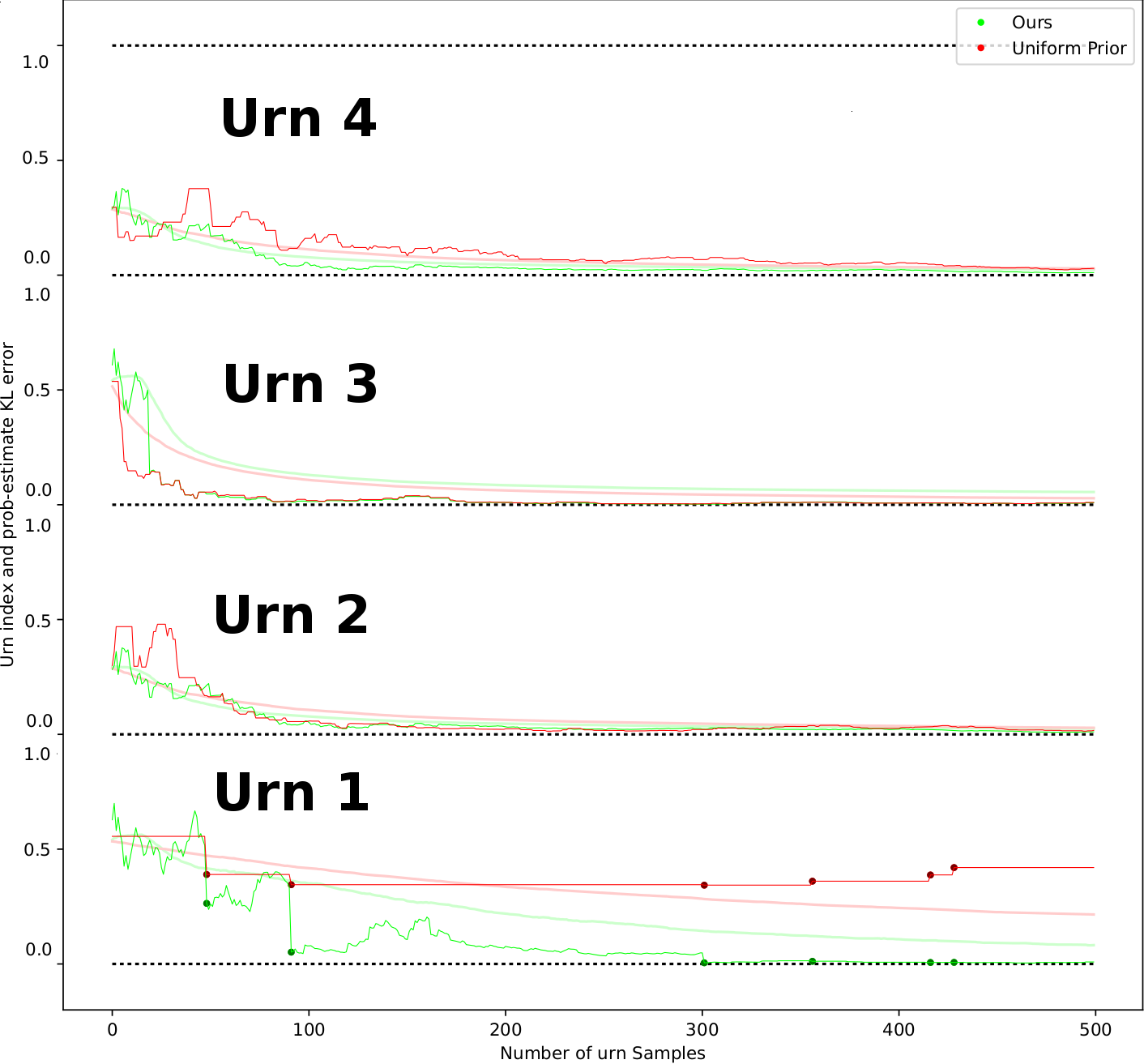}
      \label{fig:foururnsSplit}}}
  \caption{Error vs samples seen.  \subref{fig:foururns1} The total error for
    estimates of the four urns for a single run (solid lines) and averaged over 1000 runs
    (light-colored lines).  \subref{fig:foururnsSplit} The breakdown of the KL-error over the
    estimates for the four urns using just the raw tallies and using the prior that there are
    only two types of distributions.  Individual samples for Urn 1 are shown by markers at each
    sample.}
  \label{fig:dag}
\end{figure*}

To help illustrate how a simple prior might help reduce sample complexity, consider the setup where
we are given four urns $U_1, U_2, U_3, U_4$, each filled with balls of eight different colors
$c_1 \cdots c_8$.  We are given samples drawn from the urns with replacement, and we assume
that the urns are independent of each other.  In this setup, we are given samples from the urns
one at a time.  We do not get to choose which urn we sample, but we are told which urn was
sampled.  For example, we might get a sequence $\l<\l(U_1, c_5\r), \l(U_2, c_3\r), \l(U_1,
c_7\r)\r>$.  Our samples aren't uniform among the four urns: we sample $U_1$
with probability .025, while $U_2, U_3,$ and $U_4$ are each sampled with probability .325.
Our goal is to model the urns' distributions, minimizing the KL divergence
$D_{KL}\l(P||Q\r)$ from the estimated distribution $Q_i = Q\l(c_j|U_i\r)$ to the unseen true
distribution $P_i = P\l(c_j|U_i\r)$.  If we assume a uniform prior for each urn's distribution,
we can't do better than tallying the outcomes and taking the the expected values from
a Dirichlet distribution.  In this case, it takes thousands of samples to get a strong
estimate for the distribution of Urn 1 because it's sampled so infrequently, as
shown in Figure ~\ref{fig:foururnsSplit}.

If we are given prior knowledge that there are actually only two distributions instead of four,
then our sample complexity can be cut significantly.  That is, while draws from each of the
four urns are still independent of the other urns, we are told that each urn was filled will
balls sampled from one of two much larger urns, though we aren't told the distributions of the
balls in the larger urns, nor with which of the larger urns each of the four urns is filled.
Formally, we assume larger urns $a$ and $b$ such that $\forall_{i \in \l\{1,2,3,4\r\}} P_i \in
\l\{P_a, P_b\r\}$.  In this case, we use EM to alternatively update our estimates of the
classification of the distributions, then use these classification probabilities to update the
estimates of the underlying distributions.  Specifically, we estimate the probability that each
distribution is drawn from $a$ or $b$, $\forall_{i \in \l\{1,2,3,4\r\}, j \in \l\{a, b\r\}}
P\l(P_i = P_j | D, Q_a, Q_b\r)$, where $D$ is the data seen thus far, then use these estimates
to compute new estimates for $Q_a$ and $Q_b$.  We then use these values for $Q_a, Q_b$ to update
the classification probabilities $P\l(P_i = P_j | D, Q_a, Q_b\r)$, and so on until convergence.

We show the results of this process in Figures ~\ref{fig:dag}, plotting $D_{KL}\l(P||Q\r)$ as a
function of number of samples for estimates $Q$ for a single run.  In Figure
~\ref{fig:foururns1}, we see a significantly faster convergence for the case where we make use
of our priors over the ``raw'' or uniform prior.  We break this down into the error for the
estimates for the four urns in Figure ~\ref{fig:foururnsSplit}, where we see two sources for
this difference in the estimates.  The first source is that the model quickly concludes
(correctly) that Urns 4 and 2 have identical distributions.  Thus, it uses samples
from Urn 4 to inform the probability distribution of Urn 2 and vice versa.  In effect, it
doubles its samples for these urns, halving the number of samples needed to create its
probability estimates for them.

The other source of difference is the model's estimate for the probabilities of the rarely seen
Urn 1.  Of the first 1000 samples, only 16 are from Urn 1.  With 8 different ball colors, the
uninformed estimate is nowhere near convergence, having seen an average of only two samples per
color.  Conversely, after only two samples, our system correctly concludes that Urns 1
and 3 are drawn from the same distribution and ``transfers'' its knowledge about Urn 3 to Urn 1.
Note that the ``knowledge transfer'' goes both ways: our green line for Urn 3 dips
slightly below the red line.  This is because once our system concludes that Urns 1 and 3 have
identical distributions, it adds the paltry samples from Urn 1 to the tallies for the
distribution shared by both urns.
Another interpretation is that the system is primarily creating its model of the
probability distribution with samples from Urn 3, whereas it uses the few samples from Urn 1 to
{\em classify} Urn 1 as the same type as Urn 3.  An analogy might be made to the scenario where
knowing that ``Donald is a duck'' tells us much about Donald, but it also informs us a little
of what it means to be a duck.
Finally, note that the averaged error for ``Ours'' in Figure ~\ref{fig:foururns1} briefly increases
before decreasing.  Some insight might be gained to explain this by looking at the breakdowns in Figure 
~\ref{fig:foururnsSplit}.  We suspect that our model initially erroneously assigns Urn 3 to be the same distribution
as Urns 2 and 4, thus negatively transferring their tallies to Urn 3 for about 30 samples.

\section{Weaker Priors} 

In this section, we give a simple example of how we might generalize these priors to domains in
which we must simultaneously learn the structure and the probability distributions.
We present a vastly simplified version of searching for convolutional structure in images.
Conceptually, we would like to discover ``convolutions'' in images without prior knowledge of
which pixels (or vector indices) are next to which, or even that we're dealing with 2D grid,
but just given the prior knowledge that there is repeated structure.  In real images, we might
be given real-valued vectors of size 3,072.
We grossly simplify this to bit-vectors of size 12 to see if we can discover the repeated
structure for this case.  We are given one bit-vector at a time, and we can no longer
assume independence among the elements of the bit vectors.  As in the previous task, our task
is to model the 12-way joint distribution.  If we assume a uniform distribution for each
outcome, then without other priors, the best we can do is model the outcomes as a Dirichlet
distribution, with one bin for each of the $2^{12}$ possible outcomes.

We consider the reduction in sample complexity given by the following priors:
\begin{enumerate}[noitemsep,topsep=-\parskip] 
\item That the 12 variables form 4 independent distributions, each of 3 variables (though we
  are not told what the groups are).
\item That the 4 distributions are of only 2 types.  (This is the same as the prior in the
  previous section, with the exception that we are no longer given the groupings of the
  variables beforehand.)
\item We are given which of the 12 variables form the 4 groups, and the variables' order within
  the group.  This is equivalent to being told which of the 8 colors a 
  ball is, and from which urn.  With this knowledge, a vector of length 12 is equivalent to a
  sample from each of the four urns.  For example, if each sample consists of Boolean variables
  $\l\{V_1, \cdots, V_{12}\r\}$, this prior will break these into 4 ordered triples
  such as $\l(V_5, V_1, V_{11}\r)$, $\l(V_2, V_8, V_{11}\r)$, $\l(V_4, V_7, V_{3}\r)$, $\l(V_{10},
  V_6, V_9\r)$.  So if $V_5=1$, $V_1=1$, and $V_{11}=0$, this is equivalent to
  Urn 1 being color 6 or (1, 1, 0).
\end{enumerate} 

\vspace{\parskip}

\begin{figure}[ht] 
  \centering{
    \subfigure{(a)
      \includegraphics[width=65mm]{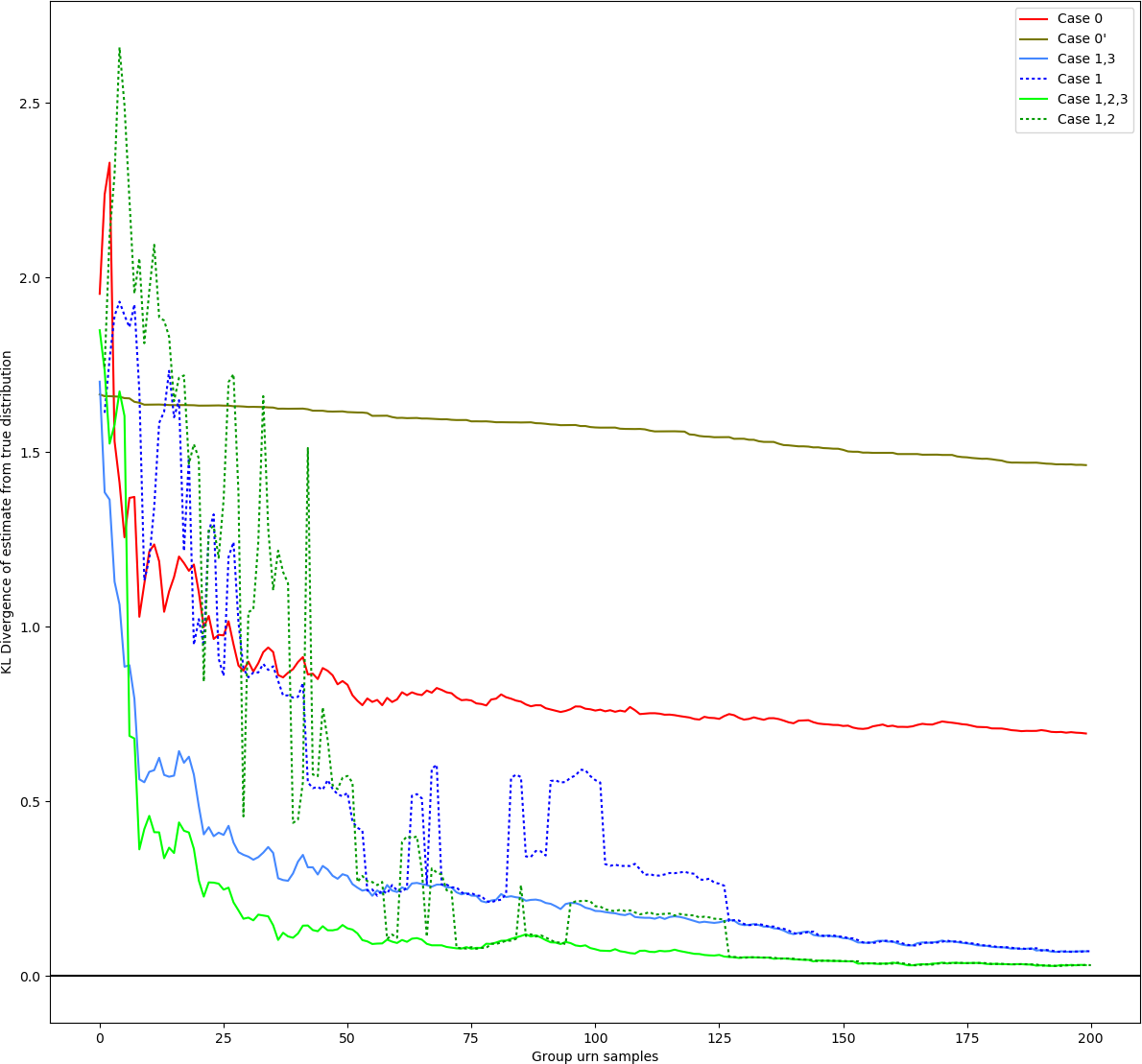}
      \label{fig:weaker}}
    \hspace{10mm}
    \subfigure{(b)
      \includegraphics[width=65mm]{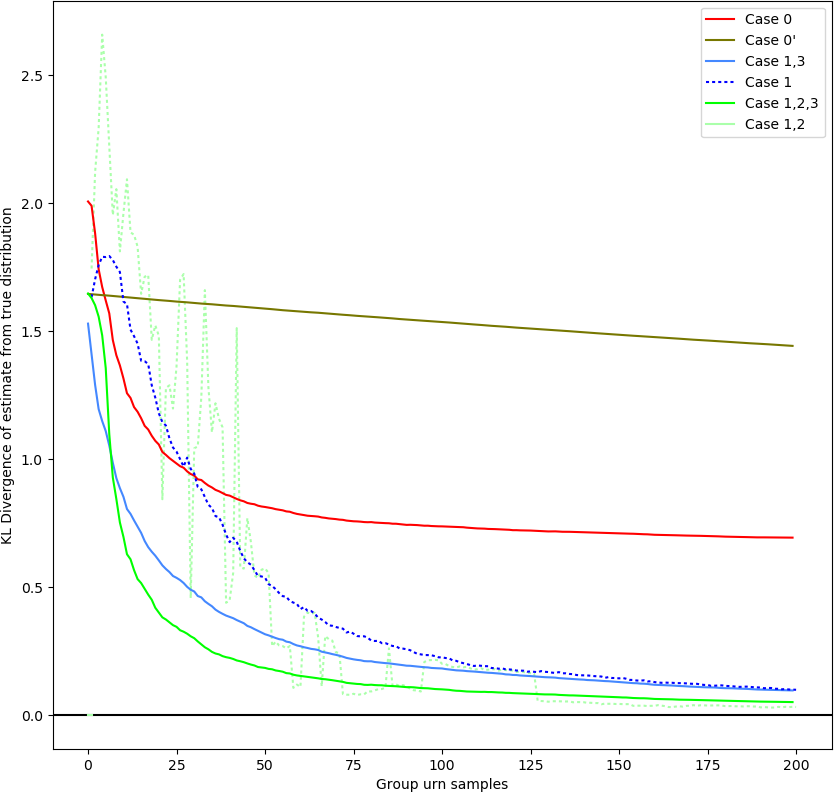}
      \label{fig:weaker2}}}
  \caption{Error vs.\ sample complexity for different priors.  \subref{fig:weaker} A single run showing that Cases 1, and 1,2 converge to Case 1,3 and 1,2,3, respectively.  \subref{fig:weaker2} The average over 100 runs for all but Case 1,2, which takes a week on 40 cores for a single run!}
  \label{fig:dagyo2}
\end{figure}

With these priors, we have the following cases, the plots of which are shown in Figure
\ref{fig:weaker}:
\begin{description}[noitemsep,topsep=-\parskip] 
\item[Case 0] We assume (incorrectly) that the 12 elements of the bit-vector are independent of
  each other.  Here, we model each variable using a beta distribution with a uniform prior.
  This model converges quickly, and plateaus after about 100 vectors, but is not expressive
  enough to represent the true distribution.
\item[Case 0'] We assume a uniform prior over each of the $2^{12}$ outcomes.  This model will
  eventually converge to the correct distribution, but takes many more than 200 samples to do
  so.
\item[Case 1,3] We assume priors 1 and 3 (we know that there are four distributions and we are
  told which variables comprise each distribution).  This is equivalent to the ``Raw tallies''
  plot in Figure \ref{fig:foururns1}, except that we sample all four urns at every timestep.
\item[Case 1,2,3] We assume all three priors.  This is equivalent to ``ours'' in Figure
  \ref{fig:foururns1} except that we sample all four urns at every timestep.
\item[Case 1] We assume prior 1, that there are 4 independent distributions of 3 variables, but
  we don't know which variables comprise each distribution.
\item[Case 1,2] This is is the most interesting case for our purposes.  We assume that there
  are four distributions (each of 3 variables), and that these four distributions are really
  only of two distinct types, though we're not told which variables are grouped together.
\end{description} 

\vspace{\parskip}

For Case 1,2 and Case 1, we do an exhaustive search over the possible groupings of the
variables and compute the most likely ordering, using similar techniques to the previous
section.  For example, one grouping of the 12 indices might be $(V_1, V_3, V_2)$, $(V_5,
V_{12}, V_4)$, $(V_7, V_{11}, V_6)$, and $(V_{10}, V_9, V_8)$.  Given a grouping, the problem
is equivalent to The Four Urns problem.  E.g., using the grouping above, if $(1, 1,
0)$ is color 6, and $V_{10}, V_9,$, and $V_8$ are 1, 1, and 0, respectively, this would be
equivalent to drawing a ball of color 6 from Urn 4.
Given a grouping, we can explore the different assignments.  Since there are
only 2 latent variables $a$ and $b$, each of the 4 distribution gets assigned to exactly one of
the latent variables (a or b).  E.g., we may say that ``Urn'' $(V_{10}, V_9, V_8)$ is assigned
as coming from either distribution $a$ or $b$.

This technique of searching all possible groupings is clearly intractable, taking exponential
time in the length of the vectors, but is feasible for our tiny vectors.  More explicitly,
given data $D_{1, \cdots, t}$, where $D_i \in \l\{0,1\r\}^{12}$, we search over each
permutation $perm$ of the ordered set $(1,\cdots, 12)$, and each possible group assignment
function $A(x) \rightarrow \l\{a,b\r\}$ to maximize $P\l(A, perm|D\r)$ using equations
\ref{eqn:eq1} and \ref{eqn:eq2}.  (Taking into account symmetries, we can reduce the number of
permutation and assignment pairs from $2^4 \cdot 12!$ to $\f{2^4 \cdot 12!}{2 \cdot 3!3!} =
106,444,800$.)

Let $D_i^p$ be the result of applying the permutation $perm$ to $D_i$, and for $j \in
\l\{1,2,3,4\r\}$ let $D^{p,j}_i$ be the $3j$ to $3\l(j+1\r)$ elements of $D^{p,j}_i$.  If we
assume an even prior of all permutations and assignments, we get:
\begin{equation} 
\begin{split} 
& P\l(perm, A|D\r) \propto P\l(D|perm, A\r) = P\l(D^p| A\r) =
\\ & \prod_{i, j} P\l(D^{p,j}_i| A_j\r) \approx \prod_{i, j} Q\l(D^{p,j}_i| A_j\r)
 = \prod_{i, j} Q_{A_j}\l(D^{p,j}_i\r)
\end{split} 
\label{eqn:eq1}
\end{equation} 
Here we compute the estimated probability $Q_{A_j}$ using a Dirichlet distribution with $2^3$
outcomes with tallies from our observations.  Since there are only two actual distributions, we
sum our tallies for instances that have the same assignment (where $\delta\l(x,y\r) = 1$ if $x = y$ else 0).
\begin{equation} 
\label{eqn:eq2}
Q_{A_j}\l(D^{p,j}_i\r) = \f{1 + \sum_{k,l} \delta\l(D^{p,l}_k, D^{p,j}_i\r)\delta\l(A_l, A_i\r)}{2^3 + 4\l|D\r|}
\end{equation} 

As indicated in Figure \ref{fig:weaker}, this search takes about 125 vector samples to converge
on the correct grouping, and then follows the same patterns as the curves in Figure
\ref{fig:foururns1}.  This shows that, at least in this case, sample complexity can be reduced
using only the first two priors.  Figure \ref{fig:weaker2} shows averages over 100 runs.

\section{Related Work} 

The Four Urns Problem can be framed as a constrained instance of latent Dirichlet
allocation (LDA) \cite{blei2003latent}, where there are two latent classes and each
``document'' or urn is a ``mixture'' of exactly one latent class.  To our knowledge, this
special case isn't directly addressed in the topic analysis literature, because the latent
classes are not allowed to mix.  It is more difficult to phrase the 12-bit vector problem
as an instance of topic analysis, because we are not given the documents/urns, but only
parts of documents and we must deduce how to put them together.

There is also some overlap between our setup and contextual bandits \cite{dudik2011efficient,
  zhou2016latent} in that both systems are motivated to make estimates of the processes true
distributions efficiently in terms of number of samples.  Our system differs in that its
utility is directly tied to the accuracy of its estimate (instead of the payoffs of bandits)
and in that it has no choice about which observation it will see next.

The main motivation for this work came from transfer learning \cite{pan2010survey,
  taylor2016lifelong, rusu2016progressive}, continual learning \cite{ring:1994,
  pickett2016growing, kirkpatrick2017overcoming}, lifelong learning \cite{ruvolo2013ella, ammar2015autonomous}, learning
to learn \cite{thrun2012learning, andrychowicz2016learning}, and multitask learning
\cite{luong2015multi}, which all share the idea that knowledge learned from one area can be
leveraged to learn from another area with fewer samples.  This paper contributes to these areas
by investigating a simple case and offering the insight that {\em a system can
  transfer knowledge between areas if it has some estimate of its respective certainty about
  those areas}.

\section{Conclusions and Future Work} 

We have shown an example where a few assumptions will allow a decrease in sample complexity.
%
%
This can be thought of as a simple example of in-domain transfer (e.g., transferring
knowledge between Urns 1 and 3).  We hypothesize that to transfer knowledge, one needs to have
some measure of certainty of our parameters, something that Bayesian approaches handle
naturally.  This measure can also be more implicit, such as by freezing weights that have been
trained until convergence \cite{rusu2016progressive}.

This work is a preliminary exploration that this can be done.  There are two main
directions we'd like to explore in future research.  The first is generalizing Priors 1 and 2.
We conjecture that both of these can be wrapped into a description length prior, where it is
cheaper to inherit from existing models of distributions than to create a new distribution from
scratch.  This would allow our model to search over both the number of distributions and
distribution {\em types}.  The second direction is to find heuristics to make the search for
structure tractable.

\clearpage
\bibliographystyle{icml2017}\bibliography{marcbib}
\end{document}